\documentclass[letterpaper, 10 pt, conference]{ieeeconf}  

\IEEEoverridecommandlockouts                              

\usepackage{graphics} 
\usepackage{graphicx}
\usepackage{amsmath} 
\usepackage{xcolor}
\usepackage{multirow}
\usepackage{bm}
\usepackage{makecell}
\usepackage{url}
\usepackage{float}
\usepackage{subcaption}
\usepackage{booktabs,makecell,multirow}
\usepackage{amsfonts}
\usepackage{mathrsfs}
\usepackage{cite}
\usepackage{color}
\usepackage{amsmath,amsfonts,amssymb}
\usepackage{amsfonts}
\usepackage[some]{background}

\usepackage{amsmath}
\usepackage{algorithm}
\usepackage[algo2e]{algorithm2e} 
\usepackage[noend]{algpseudocode}

\title{\LARGE \bf
Digital Twin-Driven Reinforcement Learning for Obstacle Avoidance in Robot Manipulators: A Self-Improving Online Training Framework
}

\author{Yuzhu Sun, Mien Van, Stephen McIlvanna, Nguyen Minh Nhat,  Kabirat Olayemi, Jack Close and Seán McLoone
	\thanks{Yuzhu Sun, Mien Van (corresponding author), Stephen McIlvanna, Nguyen Minh Nhat, Kabirat Olayemi, Jack Close and Seán McLoone are with the Centre for Intelligent Autonomous Manufacturing Systems, School of Electronics, Electrical Engineering and Computer Science, Queen's University Belfast, Northern Ireland, UK.
	{\tt\small ysun32, m.van, smcilvanna0, nnhat01, kolayemi01, jclose06, s.mcloone@qub.ac.uk}}%
}

\begin{document}

\maketitle
\thispagestyle{empty}
\pagestyle{empty}

\begin{abstract}

The evolution and growing automation of collaborative robots introduce more complexity and unpredictability to systems, highlighting the crucial need for robot's adaptability and flexibility to address the increasing complexities of their environment. 
In typical industrial production scenarios, robots are often required to be re-programmed when facing a more demanding task or even a few changes in workspace conditions. 
To increase productivity, efficiency and reduce human effort in the design process, this paper explores the potential of using digital twin combined with Reinforcement Learning (RL) to enable robots to generate self-improving collision-free trajectories in real time. 
The digital twin, acting as a virtual counterpart of the physical system, serves as a 'forward run' for monitoring, controlling, and optimizing the physical system in a safe and cost-effective manner. 
The physical system sends data to synchronize the digital system through the video feeds from cameras, which allows the virtual robot to update its observation and policy based on real scenarios. 
The bidirectional communication between digital and physical systems provides a promising platform for hardware-in-the-loop RL training through trial and error until the robot successfully adapts to its new environment.
The proposed online training framework is demonstrated on the Unfactory Xarm5 collaborative robot, where the robot end-effector aims to reach the target position while avoiding obstacles. The experiment suggest that proposed framework is capable of performing policy online training, and that there remains significant room for improvement.

\textbf{\emph{Keywords}---digital twin, reinforcement learning, robot manipulator, hardware-in-the-loop training, collision avoidance}

\end{abstract}

\section{INTRODUCTION}\label{sec1}
In recent decades, collaborative robots have become increasingly important in Industry 5.0 since the growing automation of modern manufacturing tends to re-introduce human workers back into the loop. Unlike traditional fully pre-programmed robots that operate independently behind safety cages, the latest generation of collaborative robots are equipped with a suite of sensors that enable them to operate alongside human workers in more complex environments. Therefore, the need for being adaptive, flexible and cost-effective is growing \cite{Malik2020ManMA}. In this context, the recent advancements in artificial intelligence (AI), such as machine learning, have demonstrated significant potential to enhance the smart automation and problem-solving capabilities of robots. In particular, similar to the natural learning process of humans, reinforcement learning enables a robot to learn the optimal behaviour through trial-and-error based on feedback from the environment \cite{RN313}. While reinforcement learning is a powerful tool, it still faces several limitations when applying it to solve real-world problems. Training an RL agent for the robot in the real world is costly, hazardous and even impossible the most of time. To tackle this challenge, digital twin technology is introduced to provide a virtual representation of the robot's working environment, allowing the robot to receive feedback through this simulated context of the physical world. Therefore, the combination of RL and digital twins has gained significant interest in robotics applications. Existing research mainly focuses on applying digital twins to generate synthetic data for training the deep learning model \cite{RN391, RN402, RN372, RN396}. However, this often leads to another research area: sim-to-real transfer \cite{8202133}\cite{RN401}. As a result, the performance of the trained model directly depends on the accuracy of the virtual representation and the extent to which sim-to-real techniques can compensate \cite{RN404}. Moreover, when robots are assigned more challenging tasks or the workspace environment changes, the digital twin must perform re-modelling and re-training to adapt to the new environment. This causes interruption during the tasks and potentially reduces efficiency and productivity.

\begin{figure}[b]
	\centering
	\includegraphics[width=0.49\textwidth]{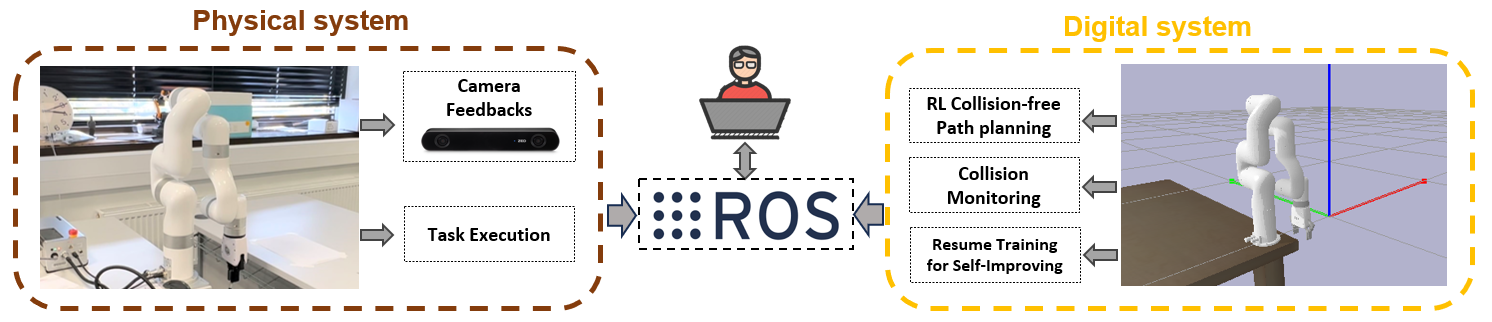}
	\caption{The structure of proposed system.}
	\label{system}
\end{figure}

In summary, while reinforcement learning combined with digital twin technology has been explored and applied to robots for specific tasks, its adaptation to collaborative robots, which requires better flexibility and problem-solving skills in uncertain environments, is still in its early stages. This paper aims to answer the question: What can a digital twin contribute to improve a robot's adaptability to an uncertain environment? Through a collision-free path planning case study, this paper develops a digital twin for a collaborative robot (i.e., the Ufactory Xarm5 robot) based on Pybullet \cite{coumans2021} and Robot Operating System (ROS), and then explores its possibilities for conducting a self-improving RL online training. In this framework, the digital twin is continuously updated in real-time using video feeds from cameras, while the RL agent constantly updates its policy when collisions or unsuccessful task attempts are detected by the digital twin's monitoring system. The structure of proposed framework is shown in Fig.\ref{system}. The main contributions of this work are as follows:

\begin{enumerate}
	\item Instead of using digital twin solely for generating synthetic data to pre-train the RL agent before task execution \cite{RN391, RN402, RN372, RN396}, we propose a self-improving online training framework conducted during the task. In this approach, the RL agent continuously updates its policy based on real-time observations through trial and error until it is able to absorb new information from the environment and successfully complete tasks that were previously deemed unachievable.
	\item According to \cite{audonnet}, compared to existing robotic simulation software, Pybullet is the most suitable software for machine learning research in terms of resource usage and idle time. The framework proposed in this paper represents the first instance of a digital twin developed based on Pybullet and the Robot Operating System (ROS) combined with a 3D ZED Depth Camera, which is expected to benefit the collaborative robotics research community.

\end{enumerate}

\section{Related Work}\label{sec2}
\subsection{Reinforcement Learning for Robotic Manipulation}
In recent years, RL algorithms have been extensively developed and applied to robot manipulators. Different from the core concepts of supervised and unsupervised learning, RL aims to maximize cumulative rewards, optimizing robots' behaviours through trial-and-error based on real-world feedback. \cite{RN412}. The existing research inculding path/motion planning \cite{RN334,RN322,RN348,RN389}, task allocation \cite{RN321}, human-robot interaction/collaboration \cite{RN312,RN386}, learning-by-demonstration \cite{RN323}, and so on. With the development of collaborative robots,  there is an increasing focus on enhancing robots' adaptability to dynamic environments. In \cite{RN387}, RL was employed to balance human avoidance and task efficiency during human-robot collaboration, utilizing a human avatar spawned and animated by skeleton data collected from a camera. Similarly, in \cite{RN313}, real-time collision avoidance and contact-rich tasks were achieved simultaneously through goal-conditioned RL, with the co-worker's hand position animated by random trajectory. It is worth noting that these approaches are offline training, as the RL agent is pre-trained before task execution. Therefore, this paper investigates a self-improving RL online training combined with digital twin technology to enhance robots' adaptability and reduce the need for human efforts in robot re-programming.

\subsection{Digital Twin}
Digital twins, often described as a virtual counterpart or digital representation of a physical entity, have garnered increasing interest over the past five years in both academia and industry \cite{RN413}. The concept of the digital twin was initially introduced by Professor Grieves of the University of Michigan \cite{grieves2014digital}, and its significance has continued growing with the development of the Internet of Things (IoT), big data analytics, sensors, robotics, simulation techniques, cloud computing and AI \cite{RN414}. Digital twins are not only valuable for designing, monitoring, and controlling physical entities but also for predicting and optimizing them in real time. These capabilities make it an ideal tool for robotics research combined with machine learning, as it offers a safe and cost-effective way for training. In robotics, the focus of digital twin applications spans various tasks such as grasping \cite{RN395}\cite{RN370}, navigation \cite{8247712}, obstacle avoidance \cite{RN417}, task allocation \cite{RN381}\cite{RN416}, and so on. It's worth mentioning that in \cite{RN410}, the path of a bridge inspection robot is online planned using a digital twin that is updated with video feeds from real robots. This hardware-in-the-loop approach enables the digital twin to be aware of real-time situations and update its strategy accordingly. Indeed, the accuracy of the digital twin may play a crucial role, as it could vary over time after the model has been trained. Inspired by the aforementioned work, this paper aims to develop a digital twin-based RL online training framework to enhance robots' adaptability to uncertain environments.

\section{Digital Twin Online Training Framework}\label{sec3}
This section provides details on the proposed online training framework, including the training algorithm, sensor perception, simulator development, and bidirectional data transmission.
\subsection{RL based Obstacle Avoidance}
\subsubsection{Markov Decision Process}
Reinforcement learning relies on an agent learning how to perform a specific goal by directly interacting with its environment through trial and error. It uses the framework of Markov decision processes (MDPs) which mathematical structures the essential features of the artificial intelligence problem \cite{RN327}. The learner to be trained is called the agent and the surroundings outside of the agent are called the environment. During the interaction, the agent receives the environment's state $S_t \in S$ and selects an action $A_t \in A$ based on the transition probability $P\left( s_{t+1}|s_t,a_t \right)$ at the time step $t$, where $S$ and $A$ is the space of states and actions. As the consequence of $A_t$, the agent receives a feedback reward $R_{t+1}\in R\subset \mathbb{R} $ and finds itself a new state $S_{t+1}$ from the environment. An MDP is thereby usually represented as the following tuple: 
\begin{equation}
	\left[ S,A,P\left( s_{t+1}|s_t,a_t \right) ,R\left( s_t,s_{t+1},a_t \right) ,\gamma \right] 
\end{equation}
The overarching objective of the agent is to maximize the cumulative rewards received from environment $G_t=\sum\nolimits_{k=0}^{\infty}{\gamma ^kr_{t+k+1}}$, where $0\leqslant \gamma \leqslant 1$ is a discount factor. A higher $\gamma$ emphasis on long-term rewards and a lower $\gamma$ tends to encourage agents to pursue immediate rewards.

\subsubsection{State}
In this paper, the objective of robot is to reach the goal (a cube) and avoidance the obstacles (boxes with different sizes). Therefore, the relevant observation for the training process including the current time step $t$, position of cube in the $x-y$ plane $pos_{goal}$, the size of obstacles $\left[h,w\right]$, the position $pos_{tcp}$ and velocity $vel_{tcp}$ of the robot end-effector and the closest points between robot $pos_{A}$ and obstacles $pos_{B}$. Thus, the state space $S$ is defined as:
\begin{equation}
	S=\left[ t, pos_{goal}, pos_{tcp}, vel_{tcp}, h, w, pos_{A}, pos_{B} \right] 
\end{equation}

\subsubsection{Action}
Choosing the appropriate action greatly influences learning performance. Typically, there are two options for defining a robot's action space. One approach is to define the action space in the joint space of the robot, which requires calculating inverse kinematics (IK) at each time step since the positions of goals and obstacles are defined in Cartesian space. However, considering that many robots on the market today come with internally built-in inverse kinematics, training a policy in Cartesian space-based policy makes it easier to transfer to other robots. Therefore, we defined the action space $A$ as:
\begin{equation}
	A=\left[ \varDelta x, \varDelta y, \varDelta z \right] 
\end{equation}
where $\varDelta x$, $\varDelta y$ and $\varDelta x$ represent the relative motion along the $x$, $y$ and $z$ axes in Pybullet world coordinates.

\subsubsection{Reward Function}
To achieve the goal of reaching the target while avoiding obstacles, the agent has two separate objectives: (i) minimizing the distance between the end-effector and the goal position, and (ii) maximizing the distance between the end-effector and obstacles. Therefore, the reward function can be designed as follows:
\begin{equation}
	r=c_1r_g+c_2r_a+c_3r_o
\end{equation}
where $r_g$ represents the reward provided by the change in distance between the end-effector and the goal position:
\begin{equation}
	r_{g_t}=d_{t-1}\left( pos_{goal_{t-1}},pos_{tcp_{t-1}} \right) - d_t\left( pos_{goal_t},pos_{tcp_t} \right) 
\end{equation}
where $d_t\left( pos_{goal_t},pos_{tcp_t} \right)$ is the distance between end-effector and obstacles at step $t$. $r_a$ denotes the penalty for the amount of effort exerted by the robot to achieve the goal, designed to assist the robot in finding an optimal path and speeding up the process:
\begin{equation}
	r_{a_t}=-\left\| \varDelta a_t \right\| 
\end{equation}
$r_o$ represents the penalty when the robot gets too close to obstacles:
\begin{equation}
	r_{o_t}\begin{cases}
		-1 \ \ \    if\,\,d\left( pos_{A_t},pos_{B_t} \right) < d_{thre}\\
		0 \ \ \ \   \    if\,\,d\left( pos_{A_t},pos_{B_t} \right) \geqslant d_{thre}\\
	\end{cases}
\end{equation}
where $d_{thre}$ is the distance threshold between robot and obstacles. $c_1$, $c_2$ and $c_3$ are weights that balance the influence of different rewards. These weights can significantly impact training, as inappropriate values may cause the learning agent to converge to a suboptimal solution rather than a globally optimal one. In this study, we utilized Proximal Policy Optimization (PPO) \cite{RN353} from Stable-Baselines3 \cite{stable-baselines3}. For brevity, detailed formulations of PPO are omitted here. 

\subsection{Object Detection and Localization}

\begin{figure}[b]
	\centering
	\includegraphics[width=0.45\textwidth]{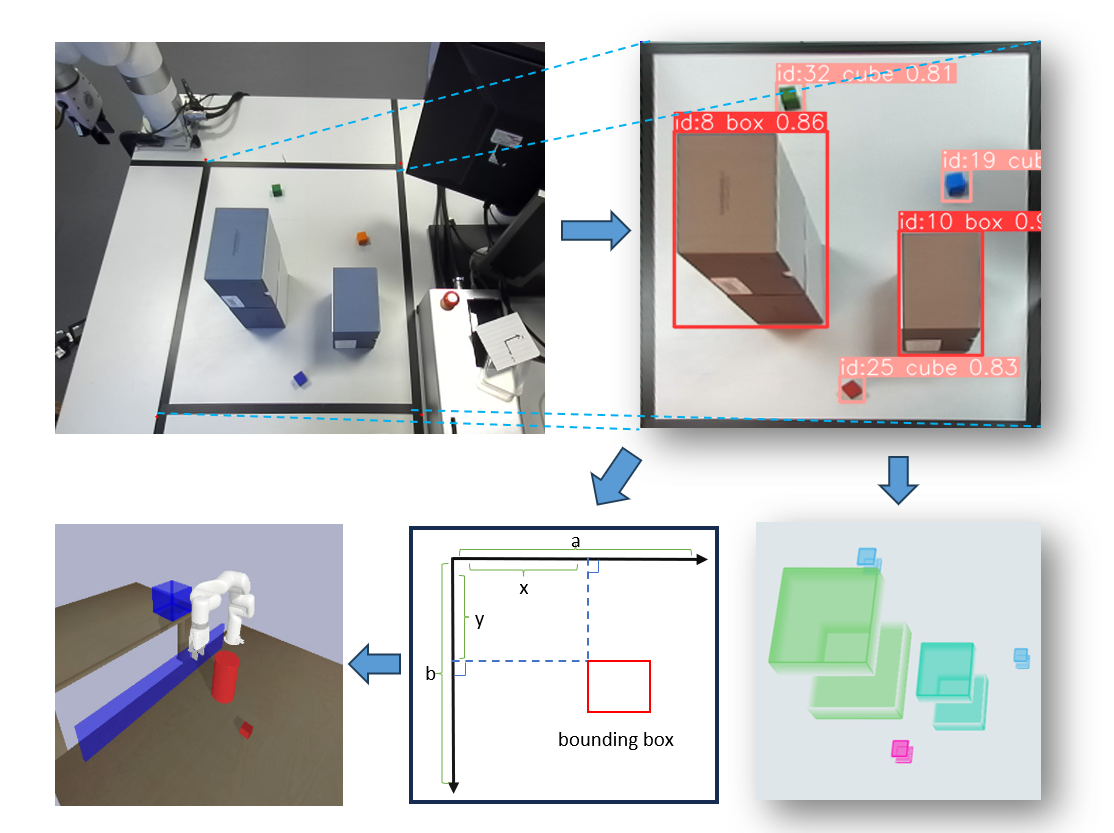}
	\caption{Object detection results.}
	\label{yolo}
\end{figure}

Transforming video feeds from cameras to 3D information involves two main steps: (i) object detection and classification of goals and obstacles, and (ii) mapping the positions of objects from real-world coordinates to Pybullet world coordinates. For object detection and classification, we utilized YOLOv8 \cite{Jocher_Ultralytics_YOLO_2023}, one of the most popular object detection algorithms known for its high accuracy and fast speed. Our object detection model was trained on a dataset of 400 images collected from cameras, achieving an mAP (mean Average Precision) of $99.4\%$. The dataset can be accessed at \cite{WinNT}. The information obtained from YOLOv8 detection includes the pixel values $x$ and $y$, as well as width $w$ and height $h$ of a 2D bounding box. By combining this information with the depth information from the ZED Stereo Camera, we are able to derive a 3D bounding box that represents the object's class, position (in pixels), and height in the real world. This integration allows for more comprehensive understanding and analysis of the objects detected in the scene. For calibrating the coordinates from pixel values to the $x$, $y$, $z$ in Pybullet world coordinate, we first collected the pixel values of four corners of the workspace (the rectangular area marked with black tape), and then apply the Perspective Transformation in OpenCV to get a transformed video so that we can get better insights into the required information that we need (as can be seen in Fig.\ref{yolo}). Finally, we calculating the ratio between the goal pixels value $x$ and the total pixels in transformed camera view $a$, and then apply exactly same ratio into the digital word according to the word coordinates in Pybullet. The mapping equation is given as follows:
\begin{equation}
\begin{cases}
	x=-0.482\left( \frac{x_p}{660} \right) +\varDelta x\\
	y=0.587\left( \frac{y_p}{540} \right) +\varDelta y\\
	z=0.025\\
\end{cases}
\end{equation}
where $x_p$ and $y_p$ are pixel values of YOLOv8 output in the transformed camera view, and 0.482 and 0.587 are the ratios mentioned previously, 660 and 540 represent the total number of pixels in the camera's horizontal and vertical dimensions, respectively, and $\varDelta x$, $\varDelta y$ denote the linear offset measured in the real world.

\subsection{Integrated Digital Twin}
The proposed digital twin is built upon Pybullet, a Python-based simulation environment utilizing the Bullet physics engine. To translate the aforementioned mathematical concepts such as agent, environment, state, and action within the simulation, we use OpenAI Gym in conjunction with Pybullet. OpenAI Gym offers a pre-defined modular framework for constructing reinforcement learning (RL) environments, which means we can easily plug in any RL algorithm to test on our custom environment as long as it get registered with OpenAI Gym. This approach facilitates the development and experimentation of RL algorithms in our simulated environment, enhancing flexibility and ease of implementation.

\begin{figure}[h]
	\centering
	\includegraphics[width=0.45\textwidth]{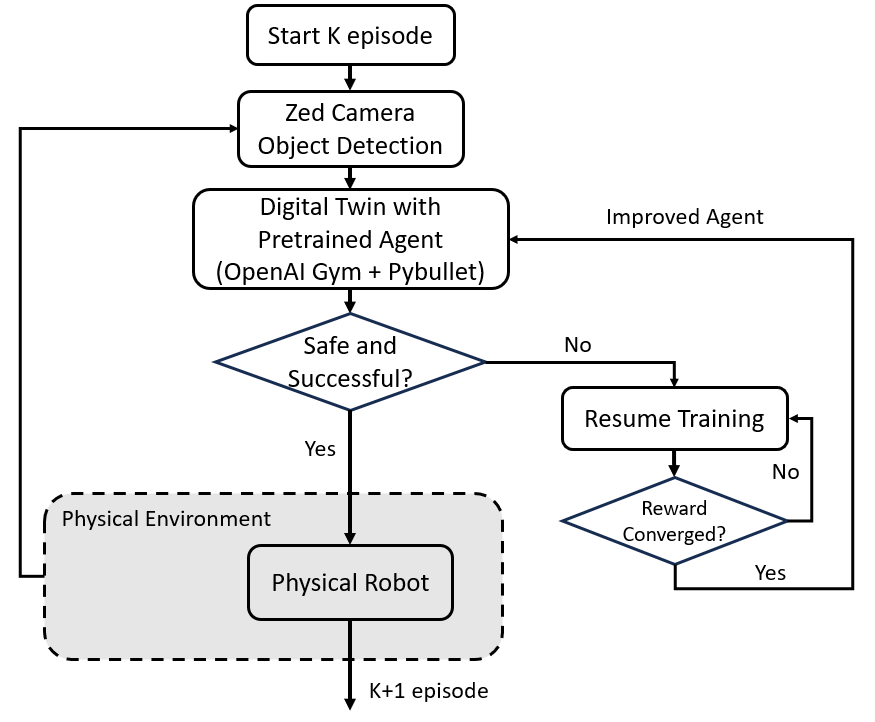}
	\caption{The flowchart of proposed framework. }
	\label{flowchart}
\end{figure}

Before initiating the proposed online training framework, the agent will be pre-trained within the digital twin environment. It will be initially trained with simpler and general objectives, such as navigating around a small obstacle. Throughout the task, bidirectional data transmission facilitated by ROS ensures synchronization between the digital and physical systems. Each component of system is programmed as a ROS node, including, ZED camera object detection node, physical robot control node, OpenAI gym environment node and resume training node. The flow chat of proposed frame is depicted in Fig. \ref{flowchart}.

Notably, the virtual robot operates one episode ahead of the real robot to predict and monitor the environment's dynamics in advance (usually taking 1-2 seconds, depending on the PC's performance). If the digital robot completes the task safely and successfully, the joint commands are transmitted from Pybullet to the physical robot. Conversely, if the digital robot encounters difficulties (can not reach the goal or have collision with obstacles), it updates its observations based on the new environment captured by the camera and resumes training based on the pre-trained model until convergence of reward is achieved again. Subsequently, the improved RL model is employed to simulate scenarios within the digital twin environment, and if deemed safe and successful, the joint poses commands are sent to the physical robot once more. The pseudocode of proposed self-improving training framework is depicted in Algorithm 1, which offers a substantial enhancement in robot adaptability, eliminating the need for repetitive reprogramming in case the environment is changed.

\begin{algorithm}
	\SetKwInOut{Input}{Input}
	\SetKwInOut{Output}{Output}
	\Input{$ t, pos_{goal}, pos_{tcp}, vel_{tcp}, h, w, pos_{A}, pos_{B} $}
	\Output{$flag_{task}$, $flag_{safe}$}
	Operate virtual robot at $k$ episode\\
	\eIf{$flag_{task} = 0$ or $flag_{safe} =  0$}
	{
		ROS publish joint configurations from Pybullet to physical robtot\\
		$k=k+1$
	}
	{
		\Repeat{Reward converged, $flag_{task} = flag_{safe} = 1$}
		{\text{Resume training} }
		\text{Publish retained joint configurations to physical robot}\\
		$k=k+1$
	}
	\caption{Self-Improving RL online training framework.}
\end{algorithm}
 
\section{Experiments and Results}\label{sec4}
\subsection{Task Description}
\begin{figure}[h]
	\centering
	\includegraphics[width=0.35\textwidth]{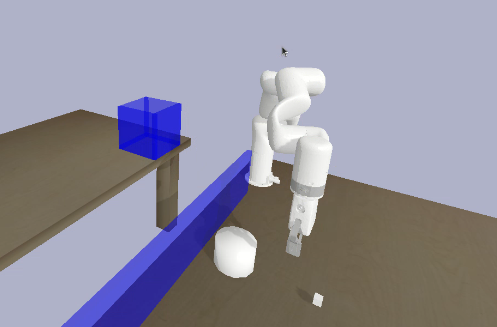}
	\caption{Pre-training robot with small obstacle.}
	\label{pretain}
\end{figure}

In this section, obstacle avoidance is selected as a case study to demonstrate the effectiveness of the proposed online training framework, as it is fundamental to the safe and efficient operation of robots in diverse environments. This experiment serves as a proof-of-concept to evaluate the efficacy of our framework. 
The objective of the experiment is for the robot to reach the goal (cubes) while simultaneously avoiding obstacles (boxes). In the case of any unsafe or unsuccessful attempts within the digital twin, the robot automatically resumes training until it achieves successful task completion.
The experiment was conducted on Ubuntu 20.04 with an Intel(R) Core(TM) i9-9960X CPU, 32GB of memory, and an NVIDIA GeForce RTX 4070Ti GPU. 
The camera used for capturing the environment is the ZED 2i Depth Camera of Stereolabs, and the robot manipulator is the Ufactory Xarm5 robot. Prior to task execution, we pre-trained an RL agent to reach the goal and avoid a small obstacle (as can be seen in Fig. \ref{pretain}), with the goal and obstacle positions randomly sampled within the observation space. The 3D model of the robot was loaded into Pybullet from the Xarm5 URDF file, and the built-in inverse kinematics (IK) function in Pybullet was utilized to translate the RL agent's actions into 5 degrees of freedom joint poses for animating the virtual robot. 

\subsection{Agent Retain Training}
After the robot is able to finish the task with small obstacles, we proceeded to replace the obstacle with a larger one. As depicted in Fig. 5(a) and Fig. 5(b), the robot collides with boxes while attempting to reach the goal, as it is assuming the obstacle was still small enough to navigate around. Detecting such collisions in real-time poses significant safety concerns and can be too late to prevent accidents, which shows the advantages of utilizing a digital twin. Pybullet provides a built-in collision detection function so that we can easily predict the collision in the digital twin. It is worth noting that we use collision detection solely to generate a flag message indicating whether the task is safe or not. For RL agent training, we employ the shortest distance function provided by Pybullet as the observation, allowing the robot to maintain a safe distance from obstacles.

\begin{figure}[htbp]
	\centering
	\begin{subfigure}{0.99\linewidth}
		\centering
		\includegraphics[width=0.9\linewidth]{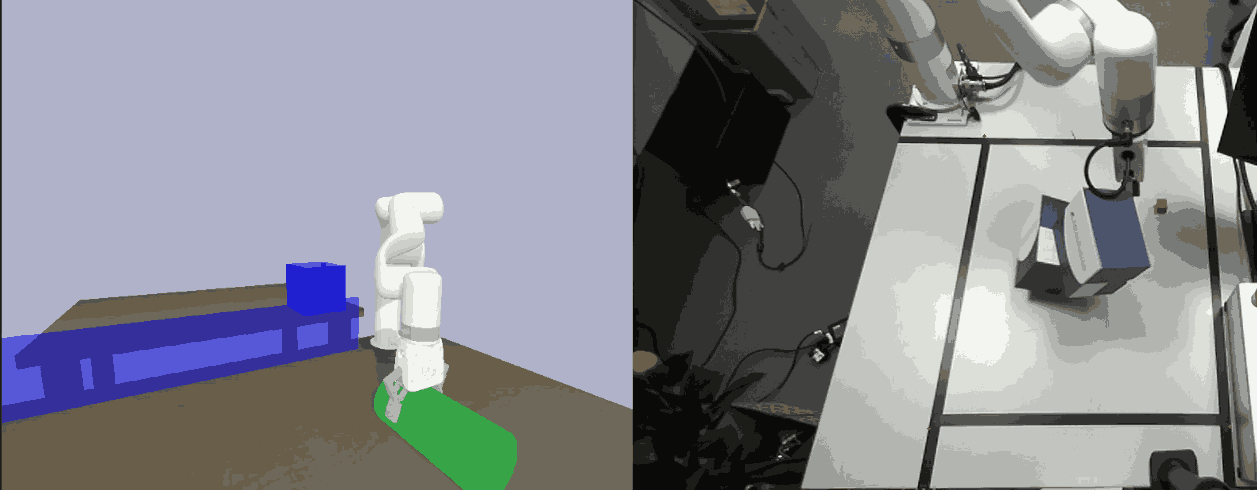}
		\caption{Digital twin }
		\label{failed1}
	\end{subfigure}
	\centering
	\begin{subfigure}{0.99\linewidth}
		\centering
		\includegraphics[width=0.9\linewidth]{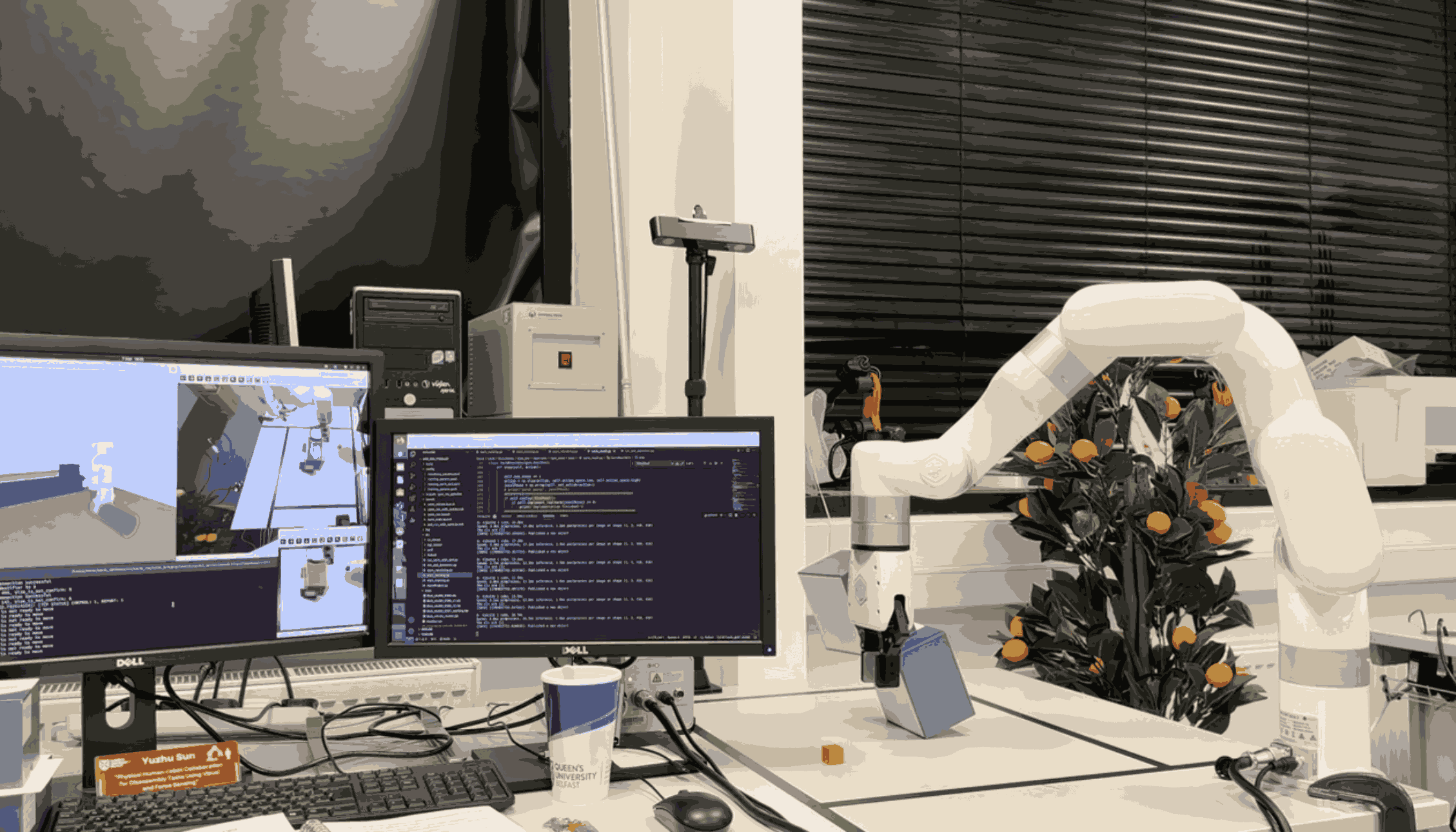}
		\caption{Physical world}
		\label{failed2}
	\end{subfigure}
\label{failed}
\caption{Robot failed to avoid higher obstacle.}
\end{figure}

The ROS launches the resume training node as soon as it receives the flag message of task failure. Fig. 6(a) and Fig. 6(b) illustrate the training reward of both the pre-trained and re-trained models. Initially, the reward of the pre-trained model starts from 0. After approximately $9.7\times 10^5$ steps of training, it reaches the optimal reward.
However, the reward sharply drops when the re-trained model starts, due to numerous collisions significantly reducing the reward.  
Subsequently, after around $1.2\times 10^4$ steps, the reward begins to rise and ultimately converges after approximately $1\times 10^5$ steps.
It is worth noting that although the pre-training is scheduled for $1\times 10^6$ steps, the optimal reward appears at $9.7\times 10^5$ steps instead of the final step, that's why the training was resumed from the model at $9.7\times 10^5$ steps. The retraining process demonstrates significant efficiency, requiring only 10\% of the time to adapt to the new environment compared to the pre-training process. Fig. \ref{retain} depicts the retraining process.
\begin{figure}[htbp]
	\centering
	\label{reward}
	\begin{subfigure}{0.99\linewidth}
		\centering
		\includegraphics[width=1\linewidth]{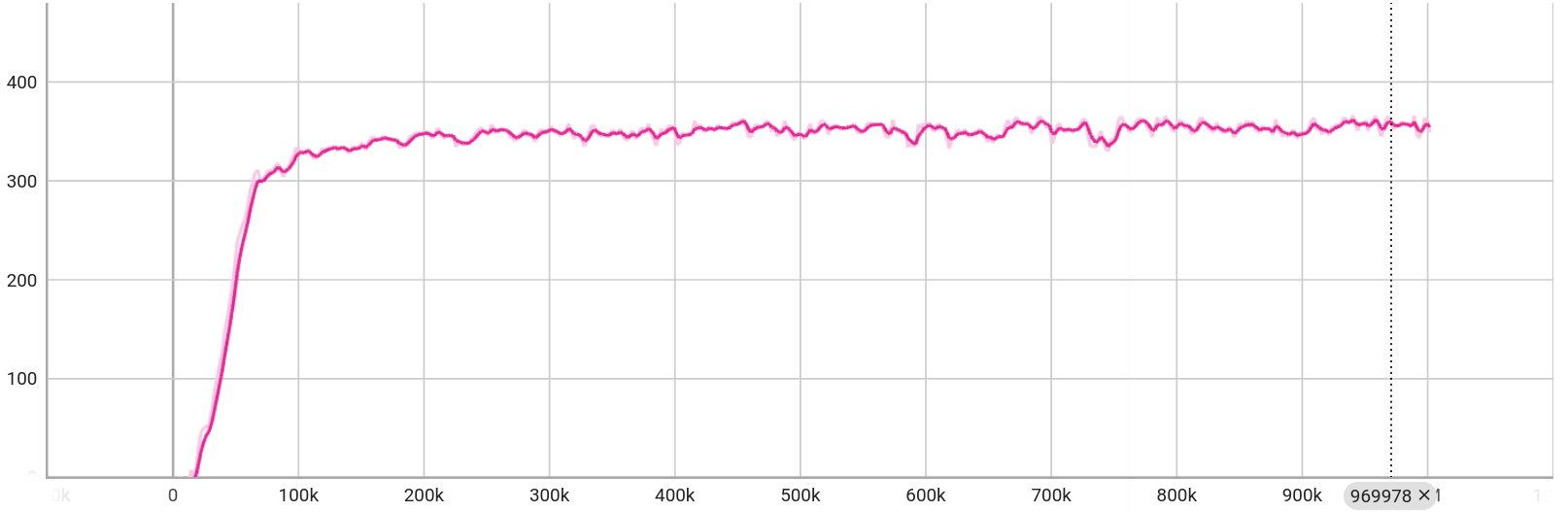}
		\caption{The reward of pre-trained model }
		\label{pre-train-reward}
	\end{subfigure}
	\centering
	\begin{subfigure}{0.99\linewidth}
		\centering
		\includegraphics[width=1\linewidth]{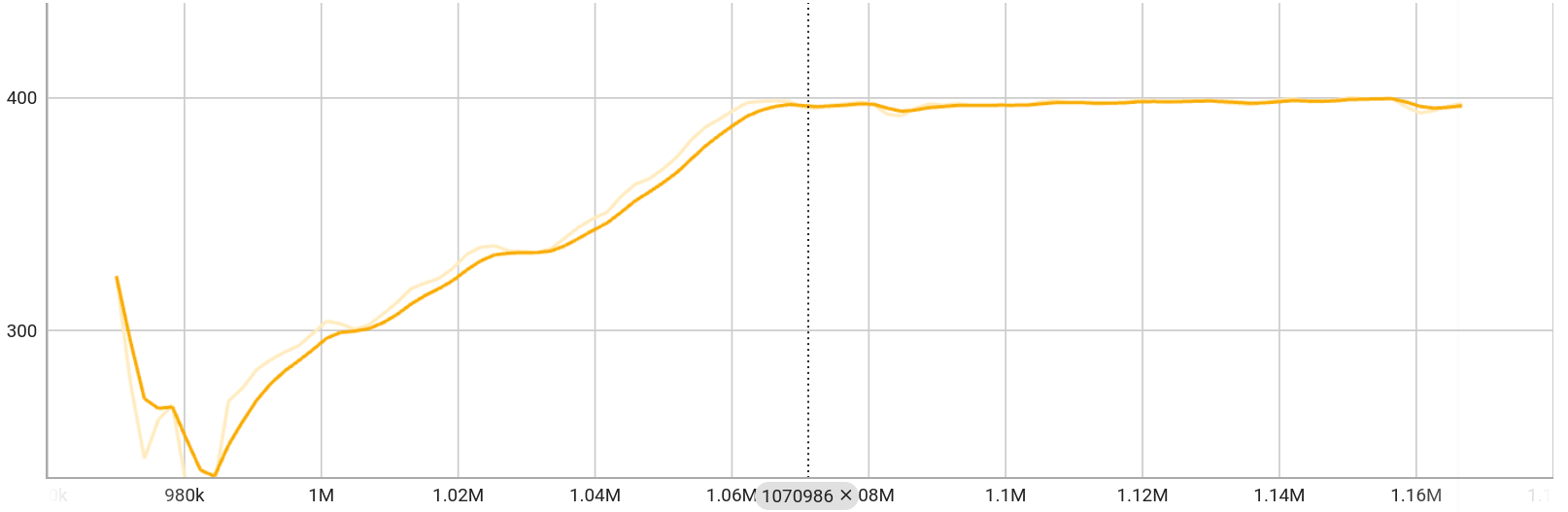}
		\caption{The reward of re-trained model}
		\label{re-train-reward}
	\end{subfigure}
\caption{Training reward.}
\end{figure}

\begin{figure}[h]
	\centering
	\includegraphics[width=0.44\textwidth]{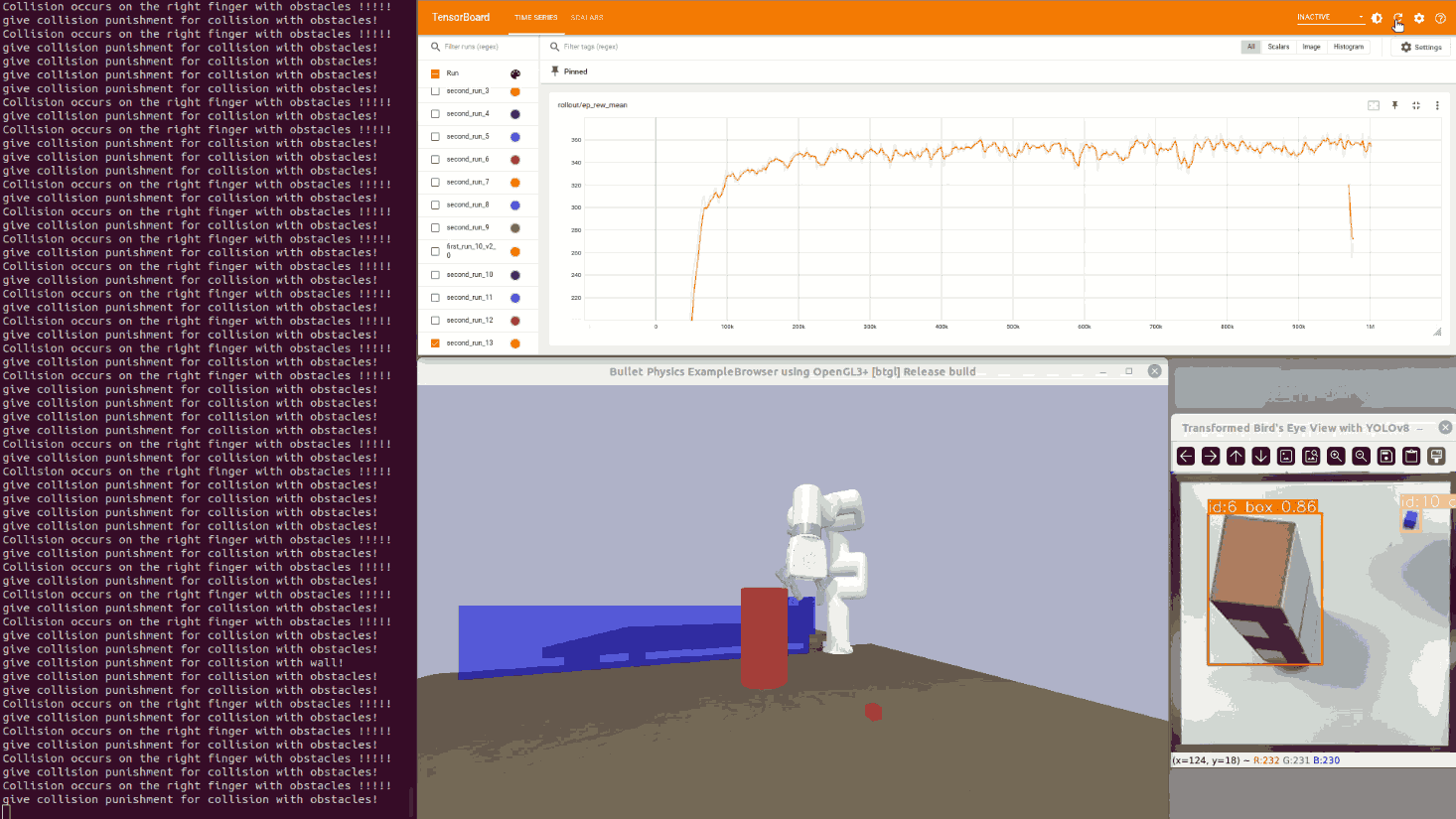}
	\caption{RL retaining}
	\label{retain}
\end{figure}

After the retraining process is completed, as demonstrated in Fig. 8(a) and Fig. 8(b), the robot autonomously avoids a higher obstacle. The video showcasing the experiment is available in \cite{mylink}.

\begin{figure}[htbp]
	\centering
	\begin{subfigure}{0.99\linewidth}
		\centering
		\includegraphics[width=0.9\linewidth]{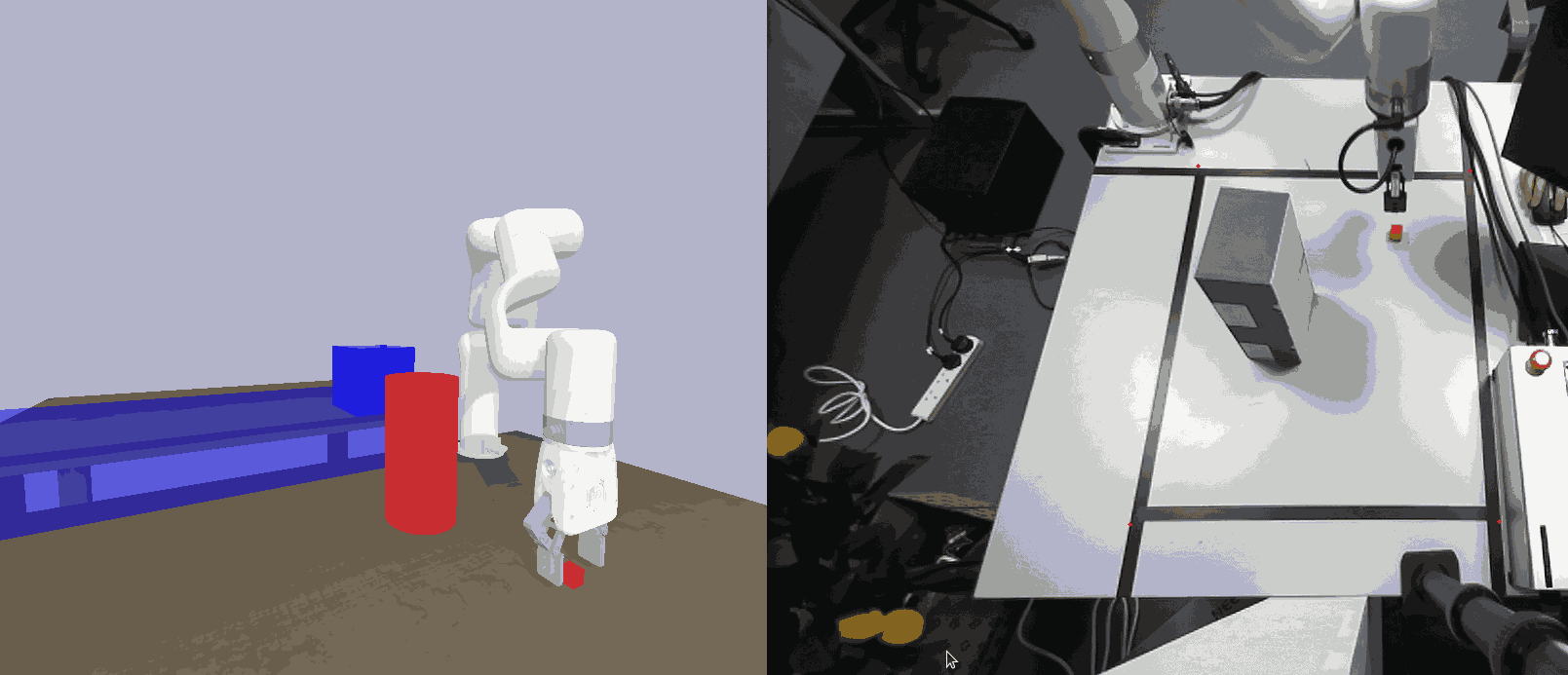}
		\caption{robot successfully avoid a higher obstacle in digital twin. }
		\label{result1}
	\end{subfigure}
	\centering
	\begin{subfigure}{0.99\linewidth}
		\centering
		\includegraphics[width=0.9\linewidth]{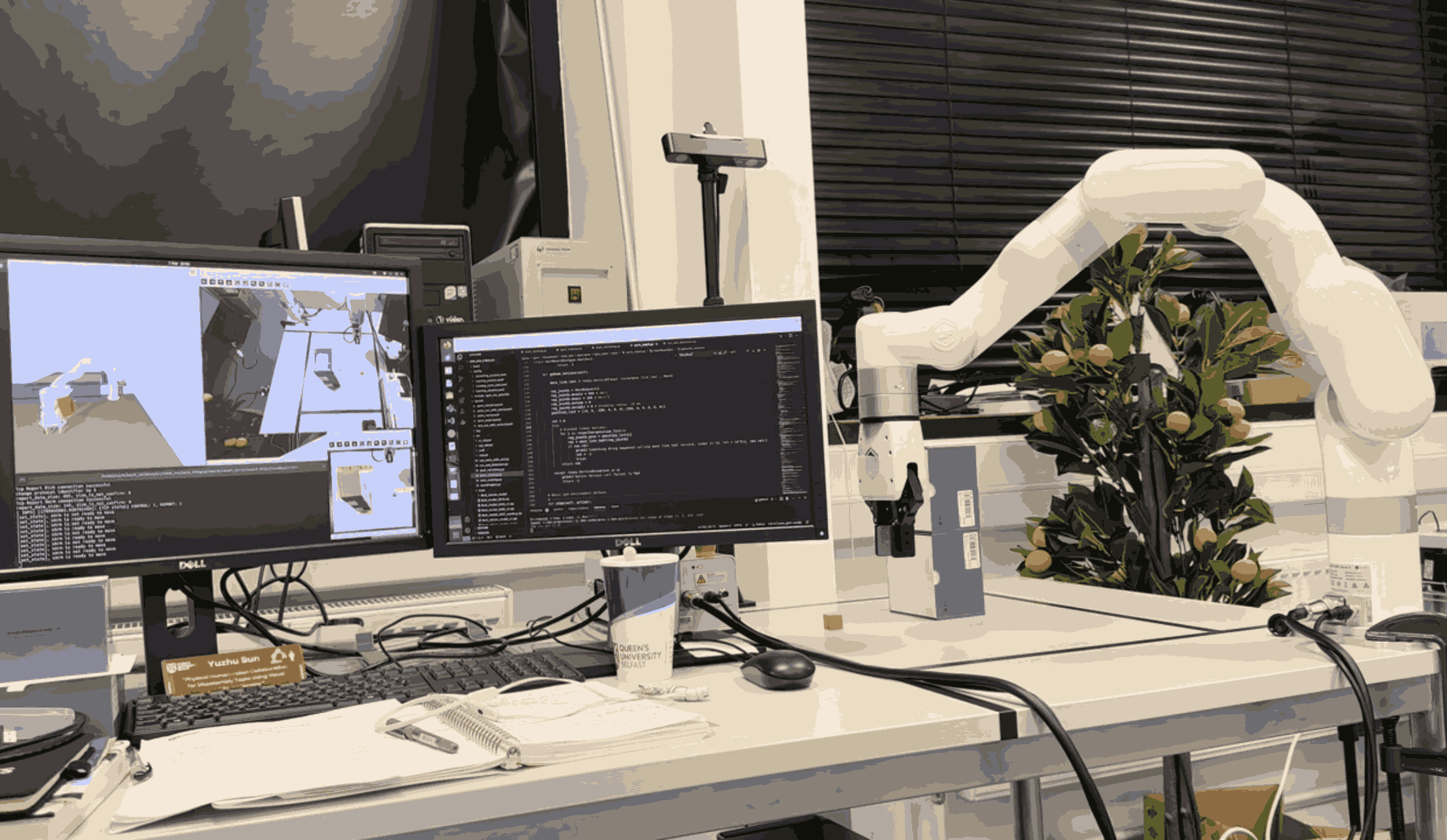}
		\caption{robot successfully avoid a higher obstacle in physical world.}
		\label{real}
	\end{subfigure}
\caption{Retain results.}
\end{figure}

\section{Conclusions}\label{sec5}
This paper presents a self-improving hardware-in-the-loop training framework that integrates digital twin with RL. Digital twin is a powerful tool for conducting machine learning research in robotics. Unlike existing researches that apply the digital twin to generate synthetic data, this paper trains RL agents directly within the digital twin environment, which continuously updates based on real-world interactions. A proof-of-concept experiment is conducted to validate the effectiveness of the proposed framework. 
Although the results show that the robot can adapt to a more challenging task than before, certain limitations exist. First, the proposed framework only works when the changes in the environment are registered inside the RL agent's observation space (in our case, the size of the obstacles). Therefore, the robot may struggle to adapt to completely unpredictable scenarios. Moreover, the camera also has its limitations in capturing the entire environmental context. Integrating multiple sensors could potentially address this limitation.







%
%
%
%

%
%
%
%
%
%
%
%

\bibliographystyle{IEEEtran}
\bibliography{References_iros}

\end{document}